\documentclass{article}
\usepackage{nips10submit_e,times}

\usepackage{amssymb}
\usepackage{amsmath}
\usepackage{amsthm}

\usepackage{subfigure}

\usepackage{algorithm}
\usepackage{algorithmic}

\usepackage{natbib}
\bibpunct{(}{)}{;}{a}{,}{,}

\usepackage{graphicx}

\title{A PAC-Bayesian Analysis of Graph Clustering and Pairwise Clustering}

\author{
Yevgeny Seldin\\
Max Planck Institute for Biological Cybernetics\\
T\"{u}bingen, Germany\\
\texttt{seldin@tuebingen.mpg.de}
}

\renewcommand{\cite}{\citep}

\begin{document} 

\newtheorem{theorem}{Theorem}

\nipsfinalcopy
\maketitle

\begin{abstract}
We formulate weighted graph clustering as a prediction problem\footnote{Pairwise clustering is equivalent to clustering of a weighted graph, where edge weights correspond to pairwise distances. Hence, from this point on, we restrict the discussion to graph clustering.}: given a subset of edge weights we analyze the ability of graph clustering to predict the remaining edge weights. This formulation enables practical and theoretical comparison of different approaches to graph clustering as well as comparison of graph clustering with other possible ways to model the graph. We adapt the PAC-Bayesian analysis of co-clustering \cite{ST08,Sel09} to derive a PAC-Bayesian generalization bound for graph clustering. The bound shows that graph clustering should optimize a trade-off between empirical data fit and the mutual information that clusters preserve on the graph nodes. A similar trade-off derived from information-theoretic considerations was already shown to produce state-of-the-art results in practice \cite{SATB05,YTS09}. This paper supports the empirical evidence by providing a better theoretical foundation, suggesting formal generalization guarantees, and offering a more accurate way to deal with finite sample issues. We derive a bound minimization algorithm and show that it provides good results in real-life problems and that the derived PAC-Bayesian bound is reasonably tight.
\end{abstract} 

\section{Introduction}

Graph clustering is an important tool in data analysis with wide variety of applications including social networks analysis, bioinformatics, image processing, and many more. As a result a multitude of different approaches to graph clustering were developed. Examples include graph cut methods \cite{SM00}, spectral clustering \cite{NJW01}, information-theoretic approaches \cite{SATB05}, to name just a few. Comparing the different approaches is usually a painful task, mainly because the goal of each of these clustering methods is formulated in terms of the solution: most clustering methods start by defining some objective functional and then minimizing it. But for a given problem how can we choose whether to apply a graph cut method, spectral clustering, or an information-theoretic approach?

In this paper we formulate weighted graph clustering as a prediction problem\footnote{Unweighted graphs can be modeled by setting the weight of present edges as 1 and absent edges as 0.}. Given a subset of edge weights we analyze the ability of graph clustering to predict the remaining edge weights. The rational behind this formulation is that if a model (not necessarily cluster-based) is able to predict with high precision all edge weights of a graph given a small subset of edge weights then it is a good model of the graph. The advantage of this formulation of graph modeling is that it is independent of a specific way chosen to model the graph and can be used to compare any two solutions, either by comparison of generalization bounds or by cross-validation. The generalization bound or cross-validation also address the finite-sample nature of the graph clustering problem and provide a clear criterion for model order selection. For very large datasets, where computational constraints can prevent considering all edges of a graph, as for example in \cite{YTS09}, the generalization bound can be used to resolve the trade-off between computational workload and precision of graph modeling.

The formulation and analysis of graph clustering presented here are based on the analysis of co-clustering suggested in \cite{ST08,Sel09}, which is reviewed briefly in section \ref{sec:co-clustering}. In section \ref{sec:graph-clustering} we adapt the analysis to derive PAC-Bayesian generalization bound for the graph clustering problem. The generalization bound depends on a trade-off between empirical fit of the cluster structure to the graph and the amount of mutual information that the clusters preserve on the graph nodes. This trade-off is related to the objective of a successful graph clustering algorithm Iclust \cite{SATB05}. We discuss this relation in section \ref{sec:related-work}. In section \ref{sec:algo} we suggest an algorithm for minimization of our bound and, finally, in section \ref{sec:applications} we present some experiments with real-world data and analyze the tightness of the bound.

\section{Review of PAC-Bayesian Analysis of Co-clustering}
\label{sec:co-clustering}

Co-clustering is a widely used method for analysis of data in the form of a matrix by simultaneous clustering of rows and columns of the matrix \cite{BDG+07}. A good illustrative example of a co-clustering problem is collaborative filtering \cite{HKTR04}. In collaborative filtering one is given a matrix of viewers by movies with ratings given by the viewers to the movies. The matrix is usually sparse and the task is to predict the missing entries. We assume that there is an unknown probability distribution $p(X_1,X_2,Y)$ over the triplets of viewer $X_1$, movie $X_2$, and rating $Y$. The goal is to build a discriminative predictor $q(Y|X_1,X_2)$ that given a viewer and movie pair will predict the expected rating $Y$. A natural form of evaluation of such predictors, no matter whether they are based on co-clustering or not, is to evaluate the expected loss $\mathbb E_{p(X_1,X_2,Y)} \mathbb E_{q(Y'|X_1,X_2)} l(Y,Y')$, where $l(Y,Y')$ is an externally provided loss function for predicting $Y'$ instead of $Y$. 

\subsection{PAC-Bayesian Analysis of Discriminative Prediction with Co-clustering}

Let ${\cal X}_1 \times .. \times {\cal X}_d \times {\cal Y}$ be a $(d+1)$-dimensional product space and assume that each ${\cal X}_i$ is categorical and its cardinality $|X_i|$ is fixed and known. We also assume that ${\cal Y}$ is finite with cardinality $|Y|$ and that the loss function $l(Y,Y')$ is bounded. In the collaborative filtering example ${\cal X}_1$ is the space of viewers, ${\cal X}_2$ is the space of movies, $d=2$, and ${\cal Y}$ is the space of ratings (e.g., on a five-star scale). The loss $l(Y,Y')$ can be, for example, an absolute loss $l(Y,Y') = |Y-Y'|$ or a quadratic loss $l(Y,Y') = (Y-Y')^2$.


We assume an existence of an unknown probability distribution $p(X_1,..,X_d,Y)$ over ${\cal X}_1 \times .. \times {\cal X}_d \times {\cal Y}$ and that a training sample of size $N$ is generated i.i.d. according to $p$. We use $\hat p(X_1,..,X_d,Y)$ to denote the empirical frequencies of $(d+1)$-tuples $\langle X_1,..,X_d,Y \rangle$ in the sample. We consider the following form of discriminative predictors:
\begin{equation}
q(Y|X_1,..,X_d) = \sum_{C_1,..,C_d} q(Y|C_1,..,C_d) \prod_{i=1}^d q(C_i|X_i).
\label{eq:qyxd}
\end{equation}
The hidden variables $C_1,..,C_d$ represent a clustering of $X_1,..,X_d$. The hidden variable $C_i$ accepts values in $\{1,..,|C_i|\}$, where $|C_i|$ is the number of clusters used along dimension $i$. The free parameters of the model \eqref{eq:qyxd} are the conditional probability distributions $q(C_i|X_i)$ which represent the probability of assigning $X_i$ to cluster $C_i$ and the conditional probability $q(Y|C_1,..,C_d)$ which represents the probability of assigning label $Y$ to cell $\langle C_1,..,C_d\rangle$ in the cluster product space. We denote the free parameters collectively by ${\cal Q} = \left \{\{q(C_i|X_i)\}_{i=1}^d, q(Y|C_1,..,C_d)\right \}$. 
We define the expected and empirical losses $L({\cal Q})$ and $\hat L({\cal Q})$ of the prediction strategy defined by ${\cal Q}$ as:
\begin{align}
L({\cal Q}) &= \mathbb E_{p(X_1,..,X_d,Y)} \mathbb E_{q(Y'|X_1,..,X_d)} l(Y,Y'),\label{eq:LQ}\\
\hat L({\cal Q}) &= \mathbb E_{\hat p(X_1,..,X_d,Y)} \mathbb E_{q(Y'|X_1,..,X_d)} l(Y,Y'),\label{eq:hatLQ}
\end{align}
where $q(Y|X_1,..,X_d)$ is defined by \eqref{eq:qyxd}. We define the mutual information $\bar I(X_i;C_i)$ corresponding to the joint distribution $\bar q(X_i;C_i) = \frac{1}{|X_i|}q(C_i|X_i)$ defined by $q(C_i|X_i)$ and a \emph{uniform} distribution over $X_i$ as:
\begin{equation}
\bar I(X_i;C_i) = \frac{1}{|X_i|} \sum_{x_i \in {\cal X}_i} \sum_{c_i=1}^{|C_i|} q(c_i|x_i) \ln \frac{q(c_i|x_i)}{\bar q(c_i)},
\label{eq:IU}
\end{equation}
where $\bar q(c_i) = \frac{1}{|X_i|} \sum_{x_i} q(c_i|x_i)$ is the marginal distribution over $C_i$. Finally, we denote the KL-divergence between two Bernoulli distributions with biases $\hat L({\cal Q})$ and $L({\cal Q})$ by
\begin{equation}
kl(\hat L({\cal Q})\|L({\cal Q})) = \hat L({\cal Q}) \ln \frac{\hat L({\cal Q})}{L({\cal Q})} + (1 - \hat L({\cal Q})) \ln \frac{1 - \hat L({\cal Q})}{1 - L({\cal Q})}.
\end{equation}

The following generalization bound for discriminative prediction with co-clustering was proved in \cite{Sel09}.
\begin{theorem}
\label{thm:grid-discriminative}
For any probability measure $p(X_1,..,X_d,Y)$ over ${\cal X}_1 \times .. \times {\cal X}_d \times {\cal Y}$ and for any loss function $l$ bounded by 1, with a probability of at least $1-\delta$ over a selection of an i.i.d. sample $S$ of size $N$ according to $p$, for all randomized classifiers ${\cal Q}=\left \{ \{q(C_i|X_i)\}_{i=1}^{d}, q(Y|C_1,..,C_d)\right \}$\textup{:} 
\begin{equation}
kl(\hat{L}({\cal Q})\|L({\cal Q})) \leq \frac{\sum_{i=1}^d \left ( |X_i| \bar I(X_i;C_i) + |C_i| \ln |X_i| \right ) + \left ( \prod_{i=1}^d |C_i| \right ) \ln |Y| + \frac{1}{2}\ln(4N) - \ln \delta}{N}.
\label{eq:grid-discriminative}
\end{equation}
\end{theorem}

In practice \citet{Sel09} replace \eqref{eq:grid-discriminative} with a parameterized trade-off
\begin{equation}
{\cal F}({\cal Q}) = \beta N \hat L({\cal Q}) + \sum_{i=1}^d n_i \bar I(X_i;C_i)
\label{eq:F}
\end{equation}
and suggest an alternating projection algorithm for finding a local minimum of ${\cal F}({\cal Q})$ (for a fixed $\beta$). Bound \eqref{eq:grid-discriminative} is minimized by applying a linear search over $\beta$ and substituting $\hat L({\cal Q})$ and $\bar I(X_i;C_i)$ obtained from optimization of ${\cal F}({\cal Q})$ back into \eqref{eq:grid-discriminative}. Alternatively, the value of $\beta$ can be tuned by cross-validation. This algorithm achieved state-of-the-art performance on the MovieLens collaborative filtering dataset. Below we adapt this analysis and algorithm to the graph clustering problem.

\section{Formulation and Analysis of Graph Clustering}
\label{sec:graph-clustering}

\subsection{Graph Clustering as a Prediction Problem}

Assume that ${\cal X}$ is a space of $|X|$ nodes and denote by $w_{ij}$ the weight of an edge connecting nodes $i$ and $j$.\footnote{All the results can be straightforwardly extended to hyper-graphs.} We assume that the weights $w_{ij}$ are generated according to an unknown probability distribution $p(W|X_1,X_2)$, where $X_1,X_2 \in {\cal X}$ are the edge endpoints. We further assume that we know the space of nodes ${\cal X}$ and are given a sample of size $N$ of edge weights, generated according to $p(X_1,X_2,W)$. The goal is to build a regression function $q(W|X_1,X_2)$ that will minimize the expected prediction error of the edge weights $\mathbb E_{p(X_1,X_2, W)} \mathbb E_{q(W'|X_1,X_2)} l(W,W')$ for some externally given loss function $l(W, W')$. Note that this formulation does not assume any specific form of $q(W|X_1,X_2)$ and enables comparison of all possible approaches to this problem.

\subsection{PAC-Bayesian Analysis of Graph Clustering}

In this work we analyze the generalization abilities of $q(W|X_1,X_2)$ based on clustering:
\begin{equation}
q(W|X_1,X_2) = \sum_{C_1,C_2} q(W|C_1,C_2) q(C_1|X_1) q(C_2|X_2).
\label{eq:qw}
\end{equation}
One can immediately see the relation between \eqref{eq:qw} and \eqref{eq:qyxd}. The only difference is that in \eqref{eq:qw} the nodes $X_1,X_2$ belong to the same space of nodes ${\cal X}$ and the conditional distribution $q(C|X)$ is shared for the mapping of endpoints of an edge. Let $\hat p(X_1,X_2,W)$ be the empirical distribution over edge weights. The empirical loss of a prediction strategy ${\cal Q} = \{q(C|X), q(W|C_1,C_2)\}$ corresponding to \eqref{eq:qw} can then be written as:
\begin{equation}
\hat L({\cal Q}) = \mathbb E_{\hat p(X_1,X_2, W)} \mathbb E_{q(W'|X_1,X_2)} l(W,W').
\label{eq:LQhat}
\end{equation}

The following generalization bound for graph clustering can be proved by a minor adaptation of the proof of theorem \ref{thm:grid-discriminative}.
\begin{theorem}
\label{thm:graph-clustering}
For any probability measure $p(X_1,X_2,W)$ over the space of nodes and edge weights ${\cal X}\times{\cal X} \times {\cal W}$ and for any loss function $l$ bounded by 1, with a probability of at least $1-\delta$ over a selection of an i.i.d. sample $S$ of size $N$ according to $p$, for all graph clustering models defined by ${\cal Q}=\left \{ q(C|X), q(W|C_1,C_2)\right \}$\textup{:} 
\begin{equation}
kl(\hat{L}({\cal Q})\|L({\cal Q})) \leq \frac{|X| \bar I(X;C) + |C| \ln |X| + |C|^2 \ln |W| + \frac{1}{2}\ln(4N) - \ln \delta}{N},
\label{eq:graph-clustering}
\end{equation}
where $|C|$ is the number of node clusters and $|W|$ is the number of distinct edge weights.
\end{theorem}
The limitation of working with a fixed set of allowed edge weights is resolved by weight quantization in section \ref{sec:quantization}.

Although there is no analytical expression for the inverse KL-divergence, given \eqref{eq:graph-clustering} we can easily bound $L({\cal Q})$ numerically:
\begin{align}
L({\cal Q}) &\leq kl^{-1}\left (\hat L({\cal Q}), \frac{|X| \bar I(X;C) + |C| \ln |X| + |C|^2 \ln |W| + \frac{1}{2}\ln(4N) - \ln \delta}{N}\right )\notag\\
            &= \max \left \{z : kl(\hat L({\cal Q})||z) \leq \frac{|X| \bar I(X;C) + |C| \ln |X| + |C|^2 \ln |W| + \frac{1}{2}\ln(4N) - \ln \delta}{N}\right \}.\label{eq:kl-bound}
\end{align}

Similar to the approach applied by \citet{Sel09} in co-clustering, in practice we can replace \eqref{eq:graph-clustering} with a parameterized trade-off:
\begin{equation}
{\cal G}({\cal Q}) = \beta N \hat L({\cal Q}) +  |X| \bar I(X;C)
\label{eq:FW}
\end{equation}
and tune $\beta$ either by substituting $\hat L({\cal Q})$ and $\bar I(X;C)$ resulting from a solution of \eqref{eq:FW} back into \eqref{eq:kl-bound} or via cross-validation. In section \ref{sec:algo} we suggest an algorithm for minimization of \eqref{eq:FW}.

\section{Related Work}
\label{sec:related-work}

The regularization of pairwise clustering by mutual information $\bar I(X;C)$ was already applied in practice by \citet{SATB05}. In their work they maximized a parameterized trade-off $\langle s \rangle - T \bar I(X;C)$, where $\langle s \rangle = \sum_{c} \bar q(c) \sum_{x_1,x_2} q(x_1|c) q(x_2|c) w_{x_1 x_2}$ measured average pairwise similarities within a cluster\footnote{The loss $L({\cal Q})$ is slightly more general than $\langle s \rangle$ since it also considers edges between the clusters.}. Their algorithm demonstrated superior results in cluster coherence compared to 18 other clustering methods. The regularization by mutual information was motivated by information-theoretic considerations inspired by the rate distortion theory \cite{CT91}. Namely, the authors drew a parallel between $\langle s \rangle$ and distortion and $\bar I(X;C)$ and compression rate of a clustering algorithm. Further, \citet{YTS09} showed that the algorithm can be run in parallel mode, where each parallel worker operates with a subset of pairwise relations at each iteration rather than all of them. Such mode of operation was motivated by inability to consider all pairwise relations in very large datasets due to computational constraints. \citet{YTS09} reported only minor empirical degradation in clustering quality, but no formal analysis and guarantees were suggested.

In light of this prior work the main contribution of our paper is not as much the introduction of the trade-off ${\cal G}({\cal Q})$ in equation \eqref{eq:FW}, but rather the formulation of graph clustering as a prediction problem and the analysis of the finite sample aspect of this problem. The experiments that follow focus on the analysis of tightness of the bound derived in section \ref{sec:graph-clustering}.

\section{An Algorithm for Graph Clustering}
\label{sec:algo}

In this section we derive an algorithm for minimization of the trade-off ${\cal G}({\cal Q})$. Unlike the co-clustering trade-off ${\cal F}({\cal Q})$ in equation \eqref{eq:F}, which is convex in $q(C_1|X_1)$ and $q(C_2|X_2)$ and thus can be minimized by alternating projections, the trade-off ${\cal G}({\cal Q})$ is not convex in $q(C|X)$. Nevertheless, we found in our experiments that alternating projections still provide good outcome in practice. Alternatively, one can apply sequential minimization techniques, as done by \citet{YTS09}. The alternating projections are much faster though and for that reason were chosen for the experiments.

The alternating projections are derived similar to alternating projection minimization in the rate distortion theory \cite{CT91}, namely by writing the Lagrangian corresponding to ${\cal G}({\cal Q})$, deriving it with respect to the free parameters and equating the derivative to zero. This procedure provides a set of self-consistent equations, which are exactly the same as those for alternating projection of ${\cal F}({\cal Q})$, hence we write the result in the Algorithm 1 box and refer the reader to \cite{Sel09} for derivation details. The only difference in our case is in the form of the derivative $\frac{\partial \hat L({\cal Q})}{\partial q(c|x)}$, which we derive next. 

\begin{algorithm}
\caption{One iteration of an alternating projection of ${\cal G}({\cal Q}) = \beta N \hat L({\cal Q}) + |X| \bar I(X;C)$.}
\label{algo:projection}
\begin{algorithmic}
\STATE \textbf{Input:} $\hat p(x_1,x_2,w)$, $q_t(C|X)$, $g_t(c_1,c_2)$, $N$, $|X|$, $|C|$, $l(w,w')$, $\beta$.
\STATE $\bar q_t(c) \gets \frac{1}{|X|} \sum_{x} q_t(c|x)$
\STATE $q_{t+1}(c|x) \gets \bar q_t(c) e^{-\beta N \frac{\partial {\hat L}({\cal Q}_t)}{\partial q(c|x)}}$
\STATE $Z_{t+1}(x) \gets \sum_{c} q_{t+1}(c|x)$
\STATE $q_{t+1}(c|x) \gets \frac{q_{t+1}(c|x)}{Z_{t+1}(x)}$
\STATE $g_{t+1}(c_1,c_2) \gets \arg \min_{w'} \sum_{w} l(w,w') \sum_{x_1,x_2} q_{t+1}(c_1|x_1) \hat p(x_1,x_2,w) q_{t+1}(c_2|x_2)$
\RETURN{$q_{t+1}(C|X), g_t(C_1,C_2)$.}
\end{algorithmic}
\end{algorithm}

For notational convenience we reformulate the problem in matrix notation. For simplicity we assume that the edge weights $w$ are sampled without repetition. This assumption usually holds in practice and it also does not affect the tightness of the analysis since the convergence rate of sampling without repetition is lower bounded by the convergence rate of sampling with repetition \cite{DEM04}. With this assumption we can represent the training data by the Hadamard (also known as Schur) entrywise matrix product $S \circ M$ (denoted by $S ~.\!* W$ in Matlab), where $S_{ij}=1$ if the edge from node $i$ to node $j$ was observed in the sample and $S_{ij}= 0$ otherwise, and $W_{ij} = w_{ij}$. In order to obtain the derivative $\frac{\partial \hat L({\cal Q})}{\partial q(c|x)}$ we have to assume a specific form of $l(w,w')$. We choose quadratic loss $l(w,w') = (w-w')^2$. The maximum likelihood reconstruction (the one that minimizes $\hat L({\cal Q})$) for the quadratic loss is a delta distribution $q(w|c_1,c_2) = \delta(w,g(c_1,c_2))$, where $g(c_1,c_2) = \arg \min_{w'} \sum_{w} l(w,w') \sum_{x_1,x_2} q(c_1|x_1) \hat p(x_1,x_2,w) q(c_2|x_2) = \sum_{x_1,x_2,w} q(c_1|x_1) w \hat p(x_1,x_2,w) q(c_2|x_2)$. This enables us to write the prediction model \eqref{eq:qw} and the loss $\hat L({\cal Q})$ in a matrix form. Let $Q$ be the matrix of $q(c|x)$ with rows indexed by cluster variables and columns indexed by node variables and $G$ be the matrix of weights predicted in the cluster product space. We denote the elements of $G$ by $g(c_1,c_2)$. The prediction model \eqref{eq:qw} can then be written as
\begin{equation}
g(x_1,x_2) = \sum_{c_1,c_2} q(c_1|x_1) g(c_1,c_2) q(c_2|x_2)
\label{eq:gx1x2}
\end{equation}
and the corresponding reconstruction matrix is $Q^T G Q$. Note that $g(x_1,x_2)$ is a function of $x_1,x_2$, which corresponds to a probability distribution $q(w|x_1,x_2)$, which is a delta function. The loss can then be written as:
\begin{equation}
\hat L({\cal Q}) = \frac{1}{N}\|S \circ (M - Q^T G Q) \|_2^2,
\label{eq:L}
\end{equation}
where $\|\cdot\|_2^2$ is the squared Frobenius norm of a matrix. The maximum likelihood $G$ is given by $G = Q (S \circ M) Q^T/N$ and the derivative $\frac{\partial \hat L({\cal Q})}{\partial q(c|x)} = 4 G^T Q (S \circ (Q^T G Q - M))/N$.

Equation \eqref{eq:L} provides an easy way to see why $\hat L({\cal Q})$ and hence ${\cal G}({\cal Q})$ are not convex in $Q$ - since $Q$ appears in forth power. Therefore, repeated iteration of alternative projections in Algorithm 1 is not guaranteed to converge (and indeed it does not). However, we found that empirically even a single iteration of Algorithm 1 achieves remarkably good results and due to simplicity of the algorithm it is easy to try multiple random initializations and obtain results comparable to those obtained by sequential optimization within much shorter time. This was the strategy followed in this paper. For large number of clusters we found it useful to anneal $\beta$ from a lower value $\beta' = 1/N$ up to the desired value in two-fold increments. At each value of $\beta$ we iterated alternating projections for 5 times and then added a small random noise to $q(c|x)$ before increasing $\beta$ by a factor of 2 until reaching the desired value.

\subsection{Correction for Edge Weight Quantization}
\label{sec:quantization}

We note that the alternating projections algorithm derived above operates with continuous weights $w$, whereas the analysis in theorem \ref{thm:graph-clustering} allows only a finite set of edge weights. If the edge weights are uniformly quantized at intervals $\Delta$, then $|W| = \frac{1}{\Delta}$ (assume that the quantization starts at $\frac{1}{2}\Delta$ and ends at $1 - \frac{1}{2}\Delta$). By rounding the continuous edge weights obtained by the alternating projections toward the closest quantization both the empirical and the expected loss are increased by at most $\Delta + \frac{1}{4}\Delta^2$. This is because quantization can shift the prediction by at most $\frac{1}{2}\Delta$ and then $l(w,w'+\frac{1}{2}\Delta) = (w - w' - \frac{1}{2}\Delta)^2 = (w-w')^2 - (w-w')\Delta + \frac{1}{4}\Delta^2 \leq l(w,w') + \Delta + \frac{1}{4}\Delta^2$, where the last inequality follows from the assumption that the loss $l(w,w')$ is bounded by 1. Hence, for the continuous weights we have
\begin{equation}
L({\cal Q}) \leq kl^{-1}\left (\hat L({\cal Q}) + \Delta + \frac{\Delta^2}{4}, \frac{|X| \bar I(X;C) + |C| \ln |X| - |C|^2 \ln \Delta + \frac{1}{2}\ln\frac{4N}{\delta^2}}{N}\right) + \Delta + \frac{\Delta^2}{4}.
\label{eq:Lwbound}
\end{equation}
As a rule of thumb we have taken $\Delta = 5 |C|^2 / N$, so that the contribution of $\Delta$ to the two operands of the inverse KL-divergence is approximately equivalent. In general this correction for quantization had no significant influence on the bound.

\section{Applications}
\label{sec:applications}

We evaluate the bound derived in section \ref{sec:graph-clustering} and the algorithm for its minimization from section \ref{sec:algo} on two real-life datasets used previously in \cite{YTS09}. The first dataset named ``king'' was taken from \cite{GSG02}. The graph represents a set of 1,740 DNS servers and the edge weights correspond to similarities between the servers. The similarities are negative exponents of the latencies between the servers scaled by dividing by the median value of all latencies in the data. The second dataset contained the graph of all known pairwise interactions among 5,202 Yeast proteins, downloaded on February 15, 2008 from the BioGRID web site\footnote{http://www.thebiogrid.org/downloads.php}. The edge weights were set to be 1 between interacting proteins and 0 otherwise.

\subsubsection*{King Dataset Experiments}

\begin{figure}[t]%
\centering
\vspace{-.4cm}
\subfigure[Bound and Test loss as a function of $\beta$]{\includegraphics[width=.5\textwidth]{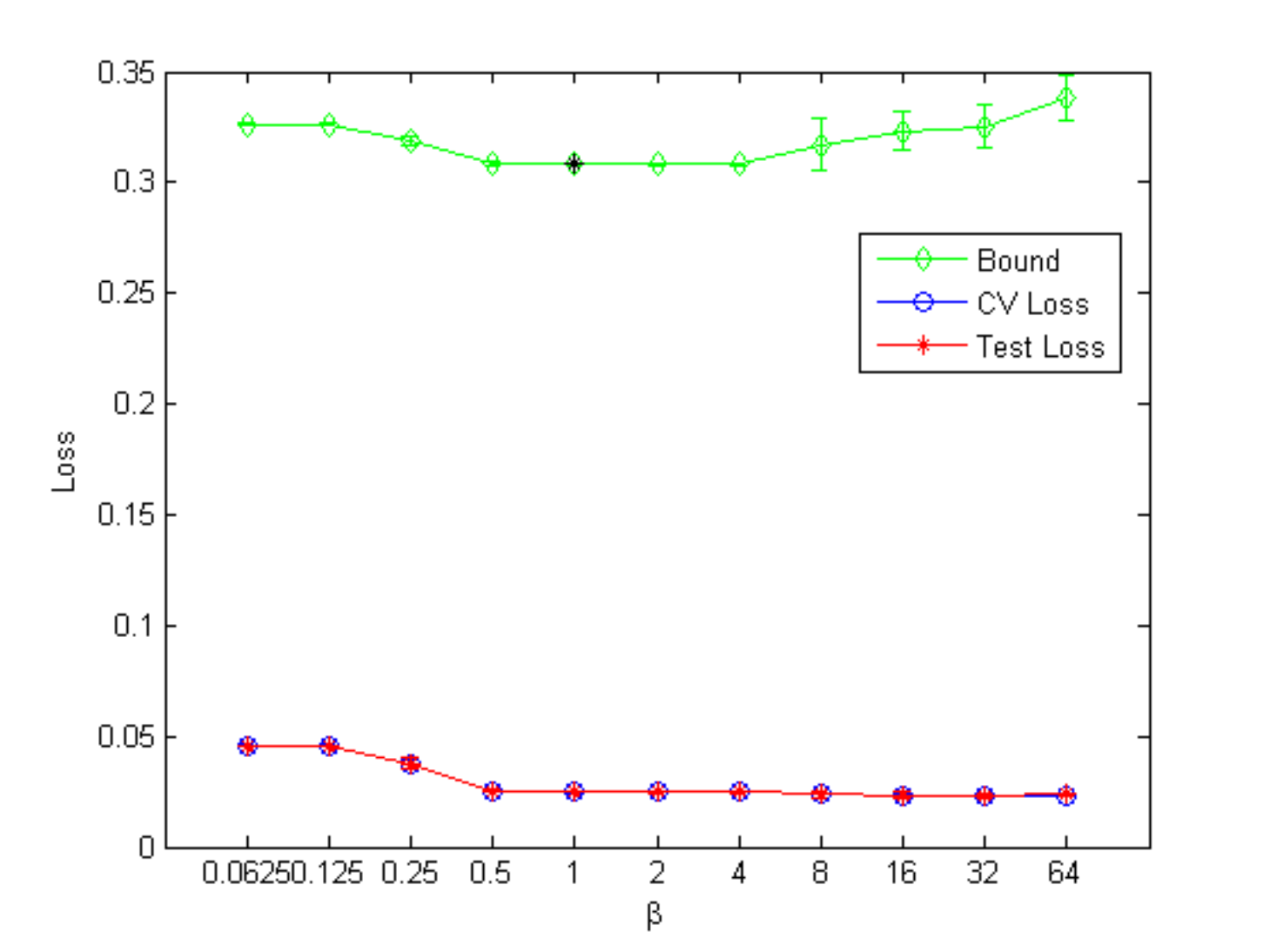}}%
\subfigure[Model Order Selection]{\includegraphics[width=.5\textwidth]{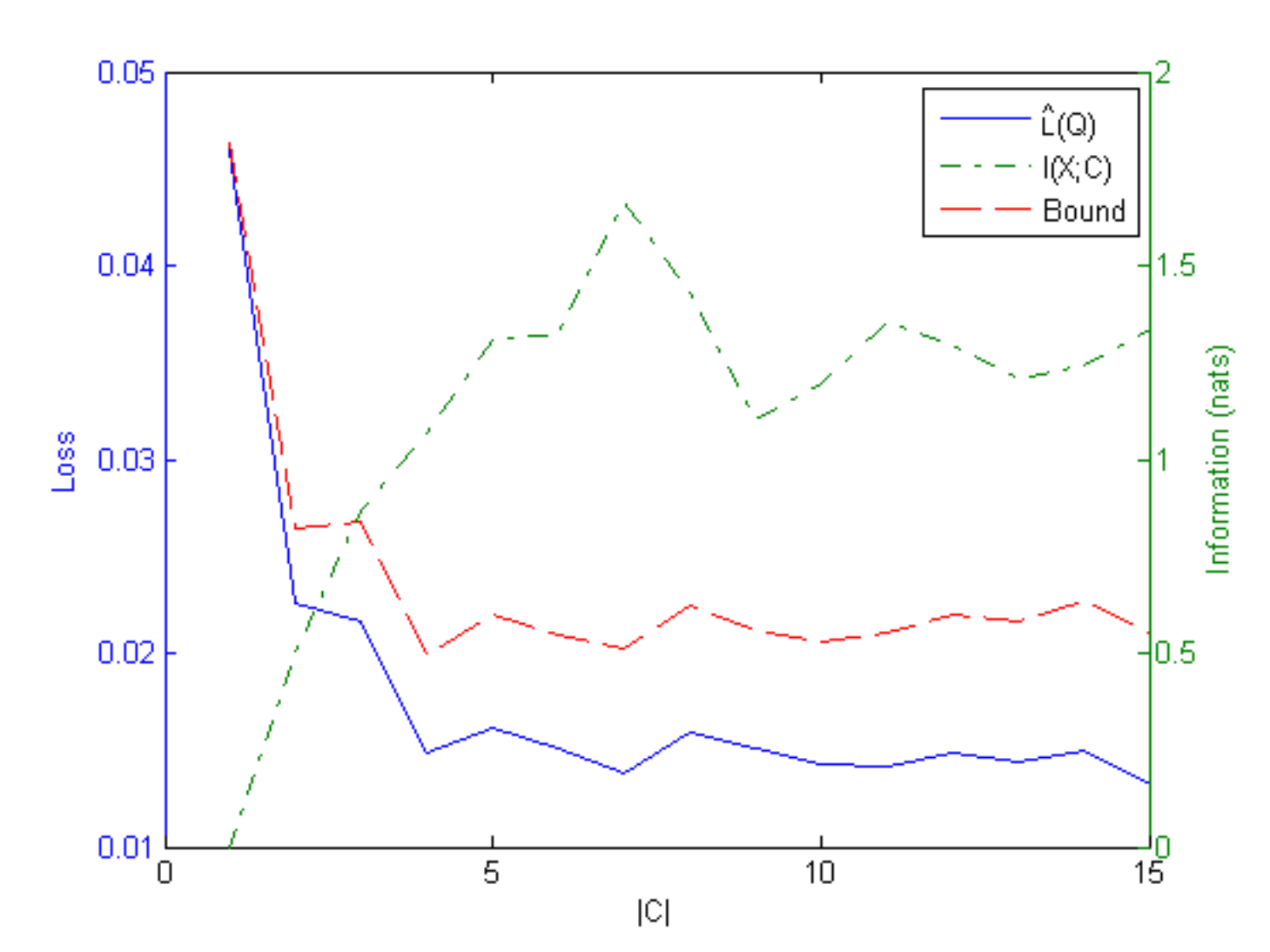}}%
\caption{{\bf King dataset experiments.} (a) Bound \eqref{eq:Lwbound}, cross-validation loss, and test loss as a function of $\beta$. Error bars indicate one standard deviation. The minimum of the bound is indicated by the black ``*''. Cross-validation follows the test loss so closely, that the curves coincide. (b) Train loss $\hat L({\cal Q})$, information $\bar I(X;C)$, and bound \eqref{eq:Lwbound} as a function of $|C|$.}%
\label{fig:king1}%
\end{figure}

In the first experiment we split the king dataset into five random train, cross-validation, and test subsets. The train set size is 103,866 edge weights, the  cross-validation set size is 25,967 edges and the test set consists of the remaining 1,383,097 edges. The size of the train set is only 3.4\% of all edges or if compared to the size of the node space the number of observed edges is $8|X|\ln|X|$. This level of sparsity is even slightly lower than the 5.3\% fraction of edges considered in each iteration of the parallel Iclust algorithm in \cite{YTS09} (the total number of edges considered in all iterations of parallel Iclust was generally larger). We cluster the graph into 41 clusters, which is the same number used by \citet{YTS09} and compare the test loss and the value of bound \eqref{eq:Lwbound} as a function of $\beta$. I.e., for each value of $\beta$ we minimize ${\cal G}({\cal Q})$ using the alternating projections algorithm and substitute the resulting $\hat L({\cal Q})$ and $\bar I(X;C)$ into \eqref{eq:Lwbound} to compute the bound. The result is shown in Figure \ref{fig:king1}.a. The bound is not perfectly tight, mainly due to the large $|C|^2 \ln |W|$ term in this case. Nevertheless, the bound is meaningful and the cross-validation loss almost coinsides with the test loss.

\begin{figure}[t]%
\centering
\vspace{-.4cm}
\subfigure[Original Dataset]{\includegraphics[width=.5\textwidth]{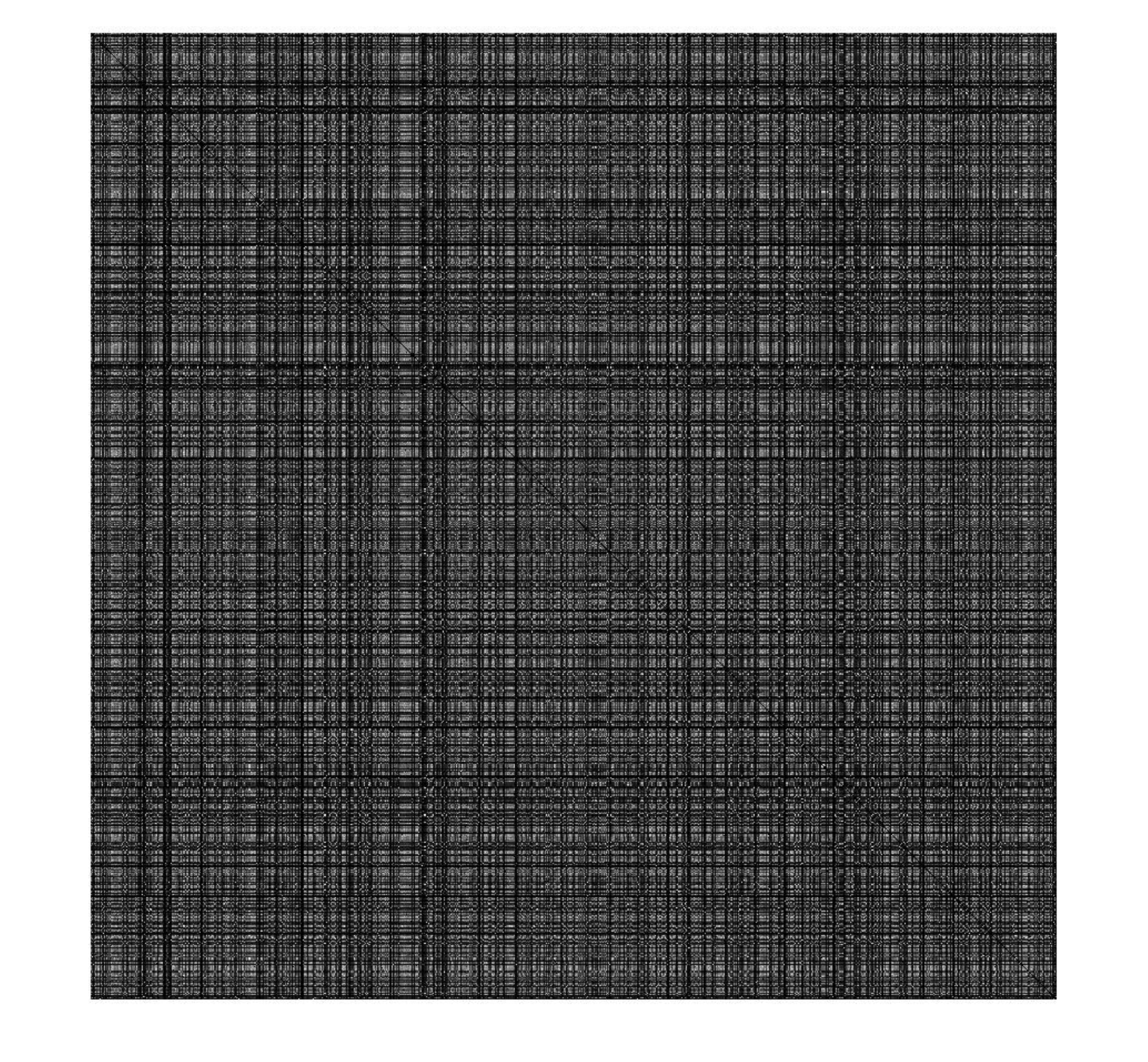}}%
\subfigure[Clustered]{\includegraphics[width=.5\textwidth]{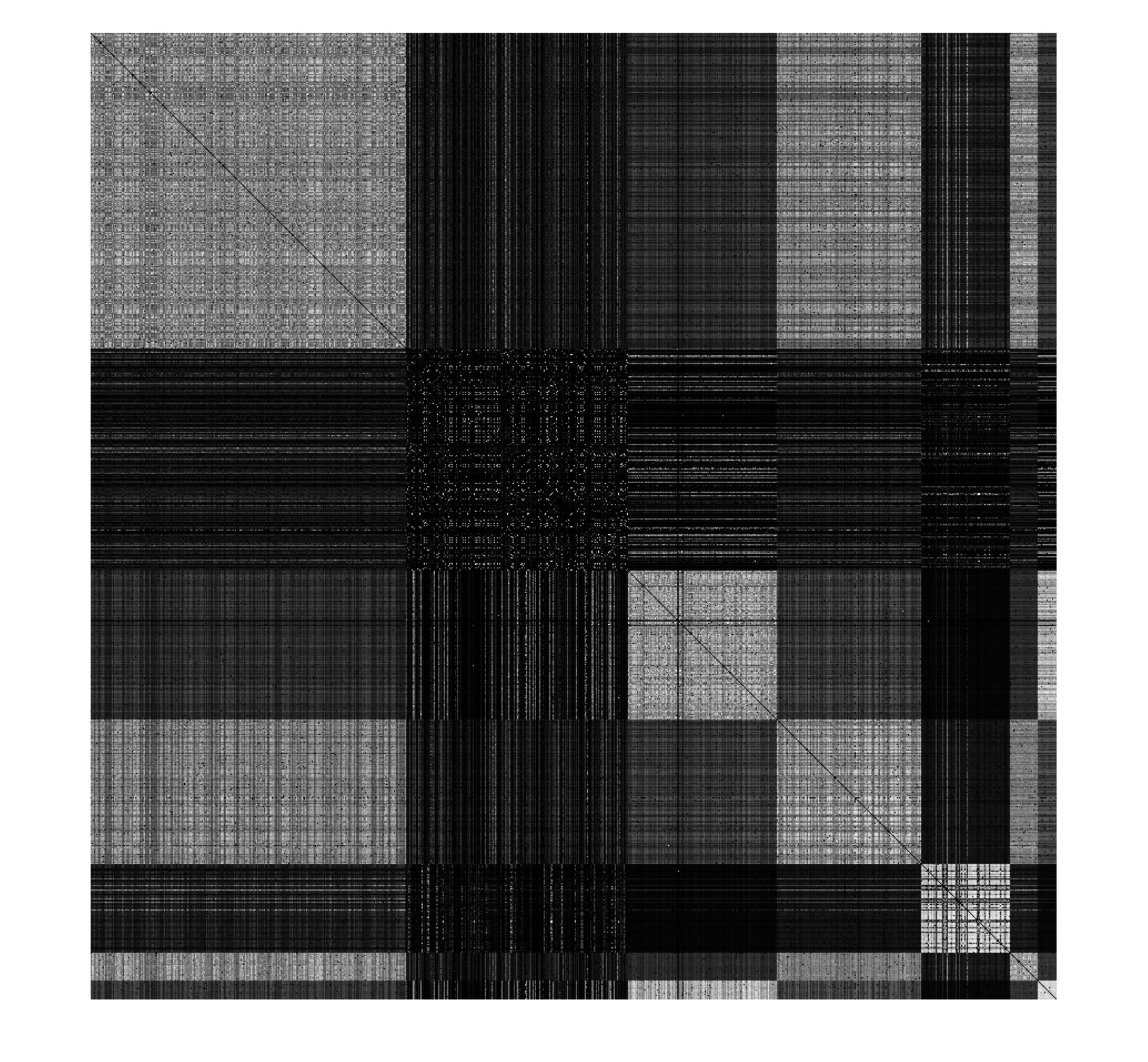}}%
\caption{{\bf Illustration of the king dataset.} (a) Original dataset. (b) Clustering into 7 clusters.}%
\label{fig:king2}%
\end{figure}

In the second experiment we consider all edges and cluster the dataset into $|C|=1,2,..,15$ clusters. (Due to symmetry every edge in this dataset appears twice, once from node $i$ to $j$ and another time from node $j$ to $i$, but in our analysis we consider only one copy of each edge.) The value of $\beta$ in the optimization trade-off ${\cal G}({\cal Q})$ was set to 1. In general by search for the optimal $\beta$ the results could be improved slightly, although as we can see from the previous experiment not considerably, so we omitted the search over $\beta$ in this experiment. The results are shown in Figure \ref{fig:king1}.b. First, we see that modeling this dataset by clustering is provably beneficial: the expected loss in predicting the weights of missing edges (would there be any) drops from 0.046 when predicting the weight with the global average to 0.02 when using four clusters and remains roughly at this level when the number of clusters is further increased. To the best of our knowledge, this is the first time when the benefit of clustering is formally proven and measured without any assumptions on the distribution that generated the edge weights (except that they were generated independently from that distribution). In this experiment there is no test set, but we can see that the bound follows the train loss pretty tightly. The mutual information preserved by the clusters on the node variables saturates at about 1.2-1.5 nats, which corresponds to effective complexity of about four clusters. Clustering of the dataset into seven clusters is illustrated in Figure \ref{fig:king2}.

\subsubsection*{Yeast Dataset Experiments}

\begin{figure}[t]%
\centering
\vspace{-.4cm}
\subfigure[Bound and Test loss as a function of $\beta$]{\includegraphics[width=.5\textwidth]{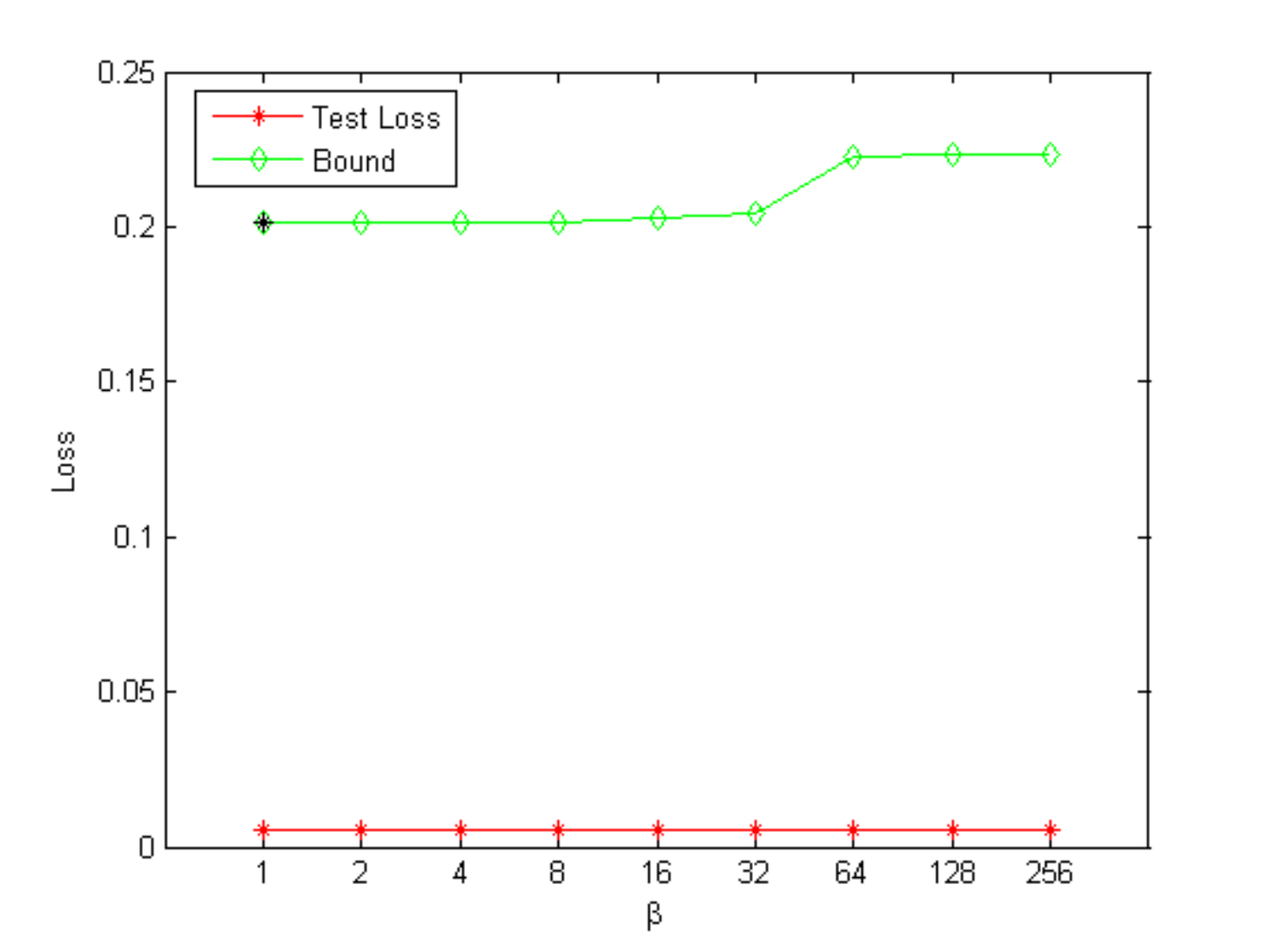}}%
\subfigure[Model Order Selection]{\includegraphics[width=.5\textwidth]{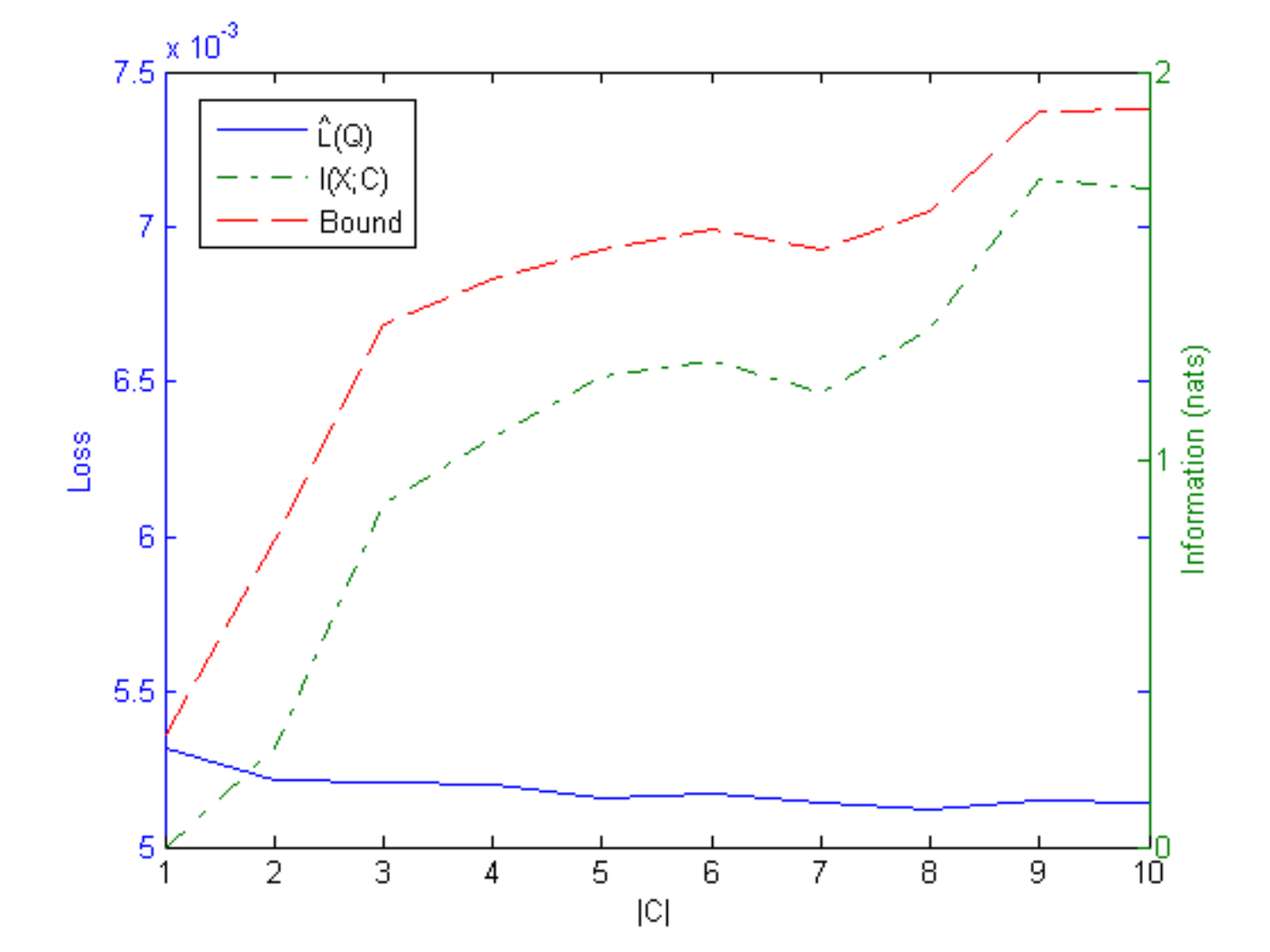}}%
\caption{{\bf Yeast dataset experiments.} (a) Bound \eqref{eq:Lwbound} and test loss as a function of $\beta$. The minimum of the bound is indicated by the black ``*''. (b) Train loss $\hat L({\cal Q})$, information $\bar I(X;C)$, and bound \eqref{eq:Lwbound} as a function of $|C|$. Note that the bound scale is on the left hand side.}%
\label{fig:yeast1}%
\end{figure}

In our first experiment we apply five random splits of the dataset into 445,125 training and 13,082,676 test edges. Training edges constitute only 3.3\% of all the edges or $10|X|\ln|X|$ if compared to the number of graph nodes. As previously, the train set sparsity is slightly lower than the 5.3\% sparsity considered in each iteration of the parallel Iclust algorithm in \cite{YTS09}. We cluster the graph into 71 clusters, which is the same number as used by \citet{YTS09}. The comparison of test loss with the value of the bound is presented in Figure \ref{fig:yeast1}.a. The bound is not perfectly tight, mainly due to the $|C|^2\ln|W|$ term, but is still meaningful.

In our second experiment we consider all edges and cluster the graph into $|C|=1,..,10$ clusters. (Symmetric edges from $i$ to $j$ and from $j$ to $i$ were considered only once.) The value of $\beta$ was set to 256. The results are shown in Figure \ref{fig:yeast1}.b. As with the king dataset experiment, the results could be slightly improved by optimizing $\beta$, however even for the large value of $\beta$ chosen the empirical loss $\hat L({\cal Q})$ exhibits very minor decrease as the number of clusters grows, hence the results would not change considerably by tuning $\beta$. Due to lower number of clusters and larger training set the bound is much tighter here than in the first yeast experiment (note that the bound y scale is on the left hand side of the graph). Unlike in the king experiment the bound tells that clustering does not help in modeling this dataset.


\section{Discussion}

We have formulated graph clustering as a prediction problem. This formulation enables direct comparison of graph clustering with any other approach to modeling the graph. By applying PAC-Bayesian analysis we have shown that graph clustering should optimize a trade-off between empirical fit of the observed graph and the mutual information that clusters preserve on the graph nodes. Prior work of \citet{SATB05} and \citet{YTS09} underscores practical benefits of such regularization. Our formulation suggests a better founded and accurate way of dealing with the finite sample nature of the graph clustering problem and tuning the trade-off between model fit and model complexity. It also suggests formal guarantees on the approximation quality. In particular such guarantees can be used for optimization of a trade-off between approximation precision and computational workload in processing of very large datasets. Our experiments show that the bound is reasonably tight for practical purposes.
\bibliography{bibliography}

\begin{thebibliography}{11}
\providecommand{\natexlab}[1]{#1}
\providecommand{\url}[1]{\texttt{#1}}
\expandafter\ifx\csname urlstyle\endcsname\relax
  \providecommand{\doi}[1]{doi: #1}\else
  \providecommand{\doi}{doi: \begingroup \urlstyle{rm}\Url}\fi

\bibitem[Banerjee et~al.(2007)Banerjee, Dhillon, Ghosh, Merugu, and
  Modha]{BDG+07}
Arindam Banerjee, Inderjit Dhillon, Joydeep Ghosh, Srujana Merugu, and
  Dhamendra Modha.
\newblock A generalized maximum entropy approach to {B}regman co-clustering and
  matrix approximation.
\newblock \emph{Journal of Machine Learning Research}, 8, 2007.

\bibitem[Cover and Thomas(1991)]{CT91}
Thomas~M. Cover and Joy~A. Thomas.
\newblock \emph{Elements of Information Theory}.
\newblock John Wiley {\&} Sons, 1991.

\bibitem[Derbeko et~al.(2004)Derbeko, El-Yaniv, and Meir]{DEM04}
Philip Derbeko, Ran El-Yaniv, and Ron Meir.
\newblock Explicit learning curves for transduction and application to
  clustering and compression algorithms.
\newblock \emph{Journal of Artificial Intelligence Research}, 22, 2004.

\bibitem[Gummadi et~al.(2002)Gummadi, Saroiu, and Gribble]{GSG02}
Krishna~P. Gummadi, Stefan Saroiu, and Steven~D. Gribble.
\newblock King: estimating latency between arbitrary internet end hosts.
\newblock In \emph{Proceedings of the 2$^{nd}$ ACM SIGCOMM Workshop on Internet
  measurement (IMW-2002)}, 2002.

\bibitem[Herlocker et~al.(2004)Herlocker, Konstan, Terveen, and Riedl]{HKTR04}
Jonathan Herlocker, Joseph Konstan, Loren Terveen, and John Riedl.
\newblock Evaluating collaborative filtering recommender systems.
\newblock \emph{ACM Transactions on Information Systems}, 22\penalty0 (1),
  2004.

\bibitem[Ng et~al.(2001)Ng, Jordan, and Weiss]{NJW01}
Andrew~Y. Ng, Michael~I. Jordan, and Yair Weiss.
\newblock On spectral clustering: Analysis and an algorithm.
\newblock In \emph{Advances in Neural Information Processing Systems (NIPS)},
  2001.

\bibitem[Seldin(2009)]{Sel09}
Yevgeny Seldin.
\newblock \emph{A PAC-Bayesian Approach to Structure Learning}.
\newblock PhD thesis, The Hebrew University of Jerusalem, 2009.

\bibitem[Seldin and Tishby(2008)]{ST08}
Yevgeny Seldin and Naftali Tishby.
\newblock Multi-classification by categorical features via clustering.
\newblock In \emph{Proceedings of the International Conference on Machine
  Learning (ICML)}, 2008.

\bibitem[Shi and Malik(2000)]{SM00}
Jianbo Shi and Jitendra Malik.
\newblock Normalized cuts and image segmentation.
\newblock \emph{IEEE Transactions on Pattern Analysis and Machine
  Intelligence}, 22\penalty0 (8), 2000.

\bibitem[Slonim et~al.(2005)Slonim, Atwal, Tracik, and Bialek]{SATB05}
Noam Slonim, Gurinder~Singh Atwal, Gasper Tracik, and William Bialek.
\newblock Information-based clustering.
\newblock \emph{Proceedings of the National Academy of Science}, 102\penalty0
  (51), 2005.

\bibitem[Yom-Tov and Slonim(2009)]{YTS09}
Elad Yom-Tov and Noam Slonim.
\newblock Parallel pairwise clustering.
\newblock In \emph{SIAM International Conference on Data Mining (SDM)}, 2009.

\end{thebibliography}

\end{document}